\documentclass[journal]{IEEEtran}

\usepackage{graphicx}
\usepackage[numbers,sort&compress]{natbib}
\usepackage{float}  %设置图片浮动位置的宏包
\usepackage{subfigure}  %插入多图时用子图显示的宏包
\usepackage{multirow} % for cmd 'multirow', 'multicolumn'
\usepackage{multicol}
\usepackage{booktabs}
\usepackage{amssymb} 
\usepackage{tabularx}
\usepackage{adjustbox}
\usepackage{array}
\usepackage{flushend}
\usepackage{xcolor}
\usepackage[T1]{fontenc}
\usepackage{aecompl}

\newcolumntype{Y}{m{0.05\textwidth}<{\centering\arraybackslash}}
\newcolumntype{Z}{m{0.2\textwidth}<{\centering\arraybackslash}}
\newcolumntype{W}{m{0.25\textwidth}<{\centering\arraybackslash}}

\begin{document}

% \title{Graph Neural Networks Enabled Efficient and Reliable Collaborative Path Planning of Multi-robot}

% \title{Intelligent Path Planning for Multi-Robot Systems: Integrating Graph Neural Networks with DHbug for Enhanced Navigation}

% \title{Collaborative Navigation for Multi-Robots in Unfamiliar Environments with Efficient Search Direction Selection}

\title{Efficient Collaborative Navigation via Perception Fusion for Multi-Robots in Unknown Environments}

\author{Qingquan Lin, 
        Weining Lu,
        Litong Meng,
        Chenxi Li,
        Bin Liang

\thanks{Corresponding Author: Dr. Weining Lu is with Beijing National Research Center for Information Science and Technology, Tsinghua University, Bejing,
China, 100084 e-mail:luwn@tsinghua.edu.cn.}% <-this % stops a space
% \thanks{J. Doe and J. Doe are with Anonymous University.}% <-this % stops a space
% \thanks{Manuscript received April 19, 2005; revised August 26, 2015.}
}

\markboth{Q. Lin et al. : Preprint submitted to IEEE }%
{Shell \MakeLowercase{\textit{et al.}}: Bare Demo of IEEEtran.cls for IEEE Journals}

\maketitle


\begin{abstract}
% Multiple mobile robots have been widely deployed in various fields, including domestic services, agriculture, and search and rescue operations.
For tasks conducted in unknown environments with efficiency requirements, real-time navigation of multi-robot systems remains challenging due to unfamiliarity with surroundings.
In this paper, we propose a novel multi-robot collaborative planning method that leverages the perception of different robots to intelligently select search directions and improve planning efficiency. Specifically, a foundational planner is employed to ensure reliable exploration towards targets in unknown environments and we introduce Graph Attention Architecture with Information Gain Weight(GIWT) to synthesizes the information from the target robot and its teammates to facilitate effective navigation around obstacles.In GIWT, after regionally encoding the relative positions of the robots along with their perceptual features, we compute the shared attention scores and incorporate the information gain obtained from neighboring robots as a supplementary weight. We design a corresponding expert data generation scheme to simulate real-world decision-making conditions for network training. 
Simulation experiments and real robot tests demonstrates that the proposed method significantly improves efficiency and enables collaborative planning for multiple robots. Our method achieves approximately $82\%$ accuracy on the expert dataset and reduces the average path length by about $8\%$ and $6\%$ across two types of tasks compared to the fundamental planner in ROS tests, and a path length reduction of over $6\%$ in real-world experiments.
\end{abstract}

% \begin{IEEEkeywords}
% Multi-robot Path Planning, Graph Neural Networks, DH-bug, Hierarchical Planning.
% \end{IEEEkeywords}

\begin{figure}[t]
\centering
\includegraphics[width=3in]{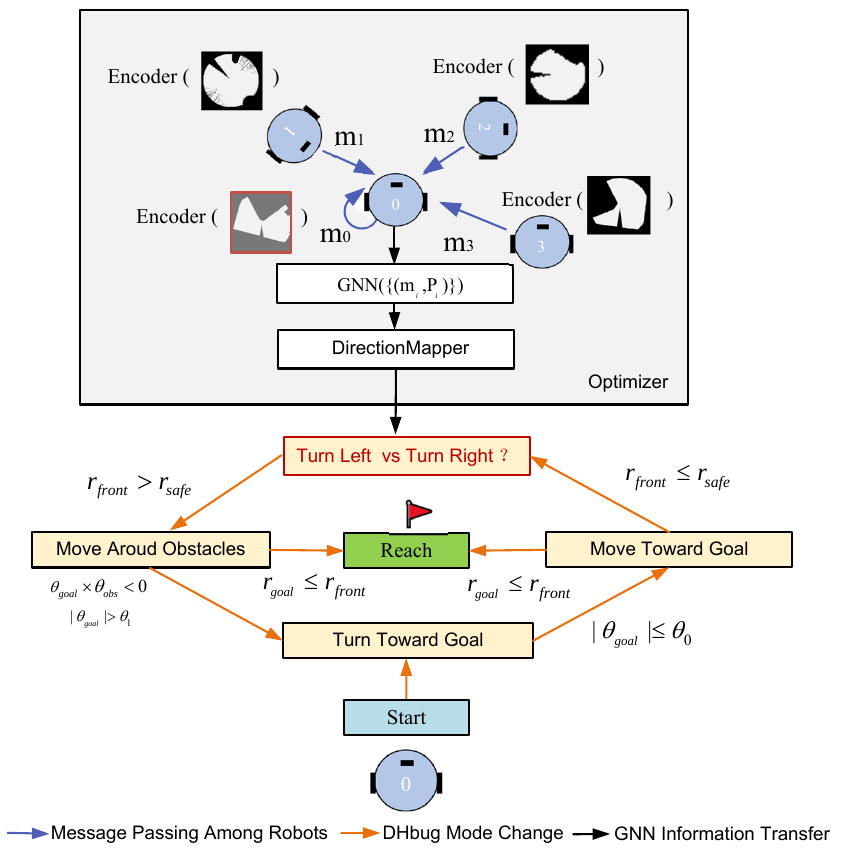}
\caption{ Method Explanation: Each robot has a local perception field of view, allowing it to detect targets, teammates, and obstacles within its perception range. When navigating toward a goal in an unknown environment, the robots utilize the DHbug algorithm to ensure they can reach their objectives when time permits. During exploration, if a robot encounters an obstacle, it employs a trained graph neural network optimizer to determine whether to turn left or right to bypass the obstacle, using its own perception and interactive information from neighboring robots.}
\label{fig:method_illustration}
\end{figure}

% ====================================================================
\section{Introduction}

\IEEEPARstart{I}{n} recent years,  multi-robots have been widely used in various domestic and outdoor services,including 
exploration\cite{sakamoto2020routing}, 
search and rescue\cite{morin2023ant}, agriculture\cite{guo2024effective} an so forth. 
For tasks conducted in unknown environments with efficiency requirements, real-time navigation of multi-robot systems remains challenging due to unfamiliarity with surroundings.  In scenarios with relatively confined spaces, one practical approach to tackle this challenge is to first create a global map of the scene using the robots' perception and localization capabilities, and then apply a global multi-robot path planning algorithm to compute optimal routes for each robot. However, although this method can produce mathematically optimal solutions based on a global map, reconstructing such a map is often time-inefficient and unnecessary, as it may not be utilized when the robots are moving away.
Consequently, how to leverage the local perception capabilities of multiple robots in unknown environments for rapid path planning has attracted widespread research attention.

Up to now, various techniques have been proposed to address this challenge, including reinforcement learning\cite{chen2023transformer},gaussian belief propagation\cite{patwardhan2022distributing}, neural networks\cite{li2020graph} and so on. 
Despite great progress have been made, there still exists different challenges for real-world applications: first grid-based methods ignore the complex distribution of obstacles under real conditions,
with an obvious gap in the sensing conditions that leads to poor performance in real-world applications
. On the other hand, the performance of the algorithm requires further improvement with regard to the convergence and the optimality. For instance, heuristic methods such as neural networks cannot guarantee the convergence of solutions, while rule-based methods lack intelligent judgment and decision-making in complex obstacle environments.

To bridge this gap and take full advantage of the distributed perception advantages of multi-robots when performing tasks, this paper proposes a hybrid multi-robot collaborative path planning method to ensure convergence and improve path efficiency in unknown environments(see Fig \ref{fig:method_illustration}). We employ the traditional Distance Histogram Bug (DHbug)\cite{DH-bug:WOS:000300267900005} algorithm as a foundational planner to ensure reliable exploration towards the target. We designed and trained a graph neural network(GNN) to provide decision support for the basic planner at critical decision points by synthesizing the local perception data from multiple robots. This method combines the convergence advantages of rule-based path planning with the capabilities of neural networks in intelligently analyzing local environments, effectively improving search efficiency while ensuring that the target is reached. The main contribution of this paper can be summarized as follows:
\begin{itemize}
    \item We propose a hybrid collaborative multi-robot path planning method, using a foundational planner to ensure reliable exploration towards the target and a trained network to provide decision support at critical points.
    \item We introduce  \textbf{G}raph \textbf{A}ttention \textbf{A}rchitecture with \textbf{I}nformation \textbf{G}ain \textbf{W}eight(\textbf{GIWT}) to efficiently synthesize the positional and perceptual data from multi-robot to facilitate effective navigation around obstacles.
    \item In order to enable the proposed method to be applied in real environments, we designed corresponding expert data generation method in The Robot Operating System(ROS) for the training of GIWT.
    \item We carried out both simulation and real robots tests to evaluate the proposed method and evaluated the key parameters.
\end{itemize}

\section{Related Work}

For decades, various techniques have been developed to solve the Multi-Robot Path Planning(MRPP) problem considering different factors like the nature of obstacles and destination,  sensing and communication conditions.

\textbf{Classic approaches} are relatively efficient at finding near-optimal solutions given a known global map. The artificial potential field method imitates the effect of force on objects to achieve path planning, where the robot moves toward the target under the combined force of the attraction of the target and the repulsive force of the obstacle\cite{zhao2020multi}.Sampling based methods like rapidly-searching random tree(RRT) use a space-filling tree to search untouched high-dimensional and non-convex space, thus generating paths \cite{shome2020drrt}. As a typical heuristic method, Conflict Based Search\cite{SHARON201540} plans optimal path for each robot using $A\star$ and then resolve conflicts in the high level using a conflict tree, ultimately planning to avoid interactions between robots.The primary limitation of classical approaches is its high computational overhead and inability to adapt to uncertainties in the environment, making it less suitable for real-time implementation.

\textbf{Bio-inspired algorithms} have gained widespread attention for their ability to effectively plan paths in complex conditions\cite{machines10090773}.
Particle Swarm Optimization is a stochastic algorithm that balances exploitation and exploration, mimicking social behavior in animals to leverage individual and group learning for both global and local searches\cite{tian2021multi}.
Genetic Algorithm follows the principle of genetics and natural selection and iteratively evolves a population of candidate paths for multi-robot planning by applying selection, crossover, and mutation operators based on their fitness evaluations\cite{nazarahari2019multi}.Ant Colony Optimization(ACO) is a metaheuristic optimization algorithm inspired by the foraging behavior of real ants, where artificial ants iteratively build solutions by depositing pheromones and making probabilistic decisions based on pheromone trails to find optimal paths\cite{liu2006path}. 
% It is worth noting that the DHbug algorithm used in this article belongs to ant colony search algorithm. 
Bio-inspired path planning methods have some problems such as slow convergence speed and local optimality, and are often combined with other methods to improve their performance.

 \textbf{Learning-based methods}, like imitation learning\cite{chen2023transformer} , reinforcement learning\cite{yang2020multi}, and recurrent neural networks\cite{lin2023multi}, have been widely studied in recent years.
In multi-robot collaborative tasks, communication and visualization among robots exhibit varying states at different times due to factors such as communication capabilities, obstacles, and communication distances, forming a dynamically changing topology. Based on this, graph neural networks have been extensively utilized in various methods\cite{wang2023hierarchical,tzes2023graph} as information fusion modules due to their exceptional capability to handle non-Euclidean data, leveraging the representation of sensory information of individual robots as nodes and their relative positions and connection status as edges in a graph data structure. Li et al.\cite{li2020graph} introduced a distributed GNN architecture to realize multi-robot decentralized planning, and then studied graph attention network with different weights to the feature of each neighbor robot\cite{li2021message}. In addition,other models like  Graph Transformer\cite{Li2023gTF}have also been studied for the MRPP problem.

Our research is closely related to bio-inspired algorithms in that we aim to develop method for each robot to reach their destination with a local field of view in uncertain environment. Unlike most learning based methods leveraging GNN as a pure feature extractor or to output the heuristic action that can be taken by robots for a single step, we utilize GNN to optimize the search direction of classic DH-bug algorithm, aiming at improve its path efficiency while maintains its advantage, convergence.

\section{ Problem Definition}
We consider a set of homogeneous mobile robots $Rs=\{R_1,R_2,\ldots,R_N \}$ which reside in a 2D environment $E$ with randomly distributed obstacles $O=\{o_1,o_2,\ldots, o_M \}$. This work addresses the Point-Goal Navigation problem where all robots are randomly deployed in an obstacle-free area $S=E\setminus O$. The objective is for all robots to collaboratively determine a series of actions to navigate towards their respective goal points, which are selected from $S$ according to task settings. 
Each robot does not have access to global position information and only possesses a local Field of View (FOV) with a fixed radius $r_{FOV}$. Within this radius, the robot can detect obstacles and other team members, and it can communicate with teammates who are within the effective communication range $r_{COM}$. A robot can detect its goal when it falls within the FOV of the robot, and when the goal is out of view, the robot can only acquire its relative direction.

Given these assumptions, the sensory information of all robots and the communication connections among robots can be represented as a geometric graph $\mathcal{G} = (\mathcal{V},\mathcal{E})$, where $\mathcal{V}=\{v_1,v_2,\ldots v_n \},v_i\in R^F$  denotes the local observations of all robots. An edge $e_{i,j}=(r_{ij},\theta_{ij}))\in \mathcal{E}$  exists if $r_{ij}<r_{COM}$, where  $r_{ij}$ represents the relative distance between robot $i$ and robot $j$, and $\theta_{ij}$ is the angle of robot $j$ relative to robot $i$. When $r<r_{COM}$, robots can communicate with each other, and there is a corresponding edge between the two nodes in $\mathcal{G}$.

\begin{figure}[]
\centering
\includegraphics[width=2.5in]{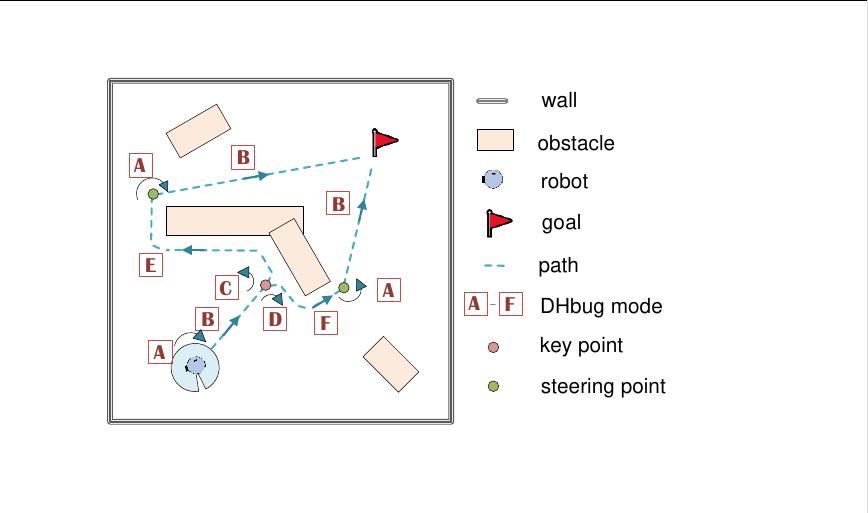}
\caption{Trajectory and Mode Switching of DHbug-Based Path Planning in Unknown Environments. When received a task, the target robot starts to turn towrd\textbf{(mode A)} and move toward the goal\textbf{(mode B)} until it encouters an obstacle at key points, where it decides to turn left\textbf{(mode C)} or turn right\textbf{(mode D)} to get around the obstacle\textbf{(mode E and F)}.In this process, if the goal and the obstacle are detected on different sides of the robot, it indicates that the obstacle has been successfully bypassed, and the robot switches to the mode A again, and so on, until the goal point is finally reached.}
\label{fig:dhbug_illustration}
\end{figure}

\section{Methodology}
\subsection{Overview of the hierarchical planning architecture }
We propose an efficient and reliable hierarchical colaborative path planning method based on graph neural networks. As illustrated in Figure \ref{fig:method_illustration}, DHbug acts as the fundamental planner that generates precise speed and angular velocity based on radar data, mathematically ensuring the convergence of the solution. When the execution time is sufficiently long, the robot will reach its accessible destination according to this base planner. However, DHbug lacks the ability to synthesize local environment data at key decision points for turning. To leverage the perceptual advantages of multiple robots, we employ graph neural networks to integrate perception data obtained from nearby neighbors and itself, intelligently selecting search directions at key points. In order to guide direction selection effectively, an expert data generation scheme and an efficient network architecture are carefully designed to fully utilize the relative positional information and perceptual data of teammates .Furthermore, we adopt a priority-based strategy to coordinate the movements of robots to prevent collisions.

\subsection{Fundamental Planner for Low-level Precise Speed Output}
We use the DHbug algorithm\cite{DH-bug:WOS:000300267900005}  as a real-time fundamental planner based on the robot's perception of the local environment and the target direction, as shown in Figure \ref{fig:dhbug_illustration}. It has been demonstrated that this algorithm can generate a convergent path solution, regardless of the complexity of the environment, as long as the goal is accessible. 

\subsubsection{\textbf{Safe Speed Calculation}}

\begin{figure}[t]
	\centering  %图片全局居中
	\subfigcapskip=-5pt %设置子图与子标题之间的距离
        \subfigure[ ]{
        \includegraphics[width=1.2in]{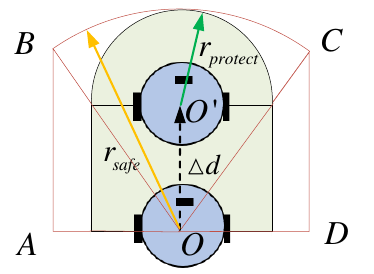}}
	\subfigure[]{
	\includegraphics[width=2in]{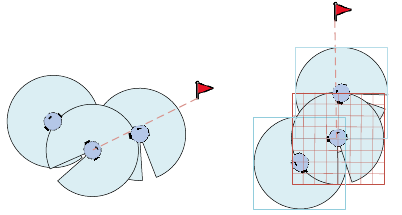}}\\
	\caption{(a) The minimum safe zone(light green) when moving forward and  its approximate calculation area(zone ABCD). (b) Relative coordinate system transformation and meshing.}
        \label{fig:safe_zone_preprocess}
\end{figure}

In consideration of the physical size of each robot, it is necessary to ensure the robots not to collide with obstacles when calculating the minimum speed in the forward direction. As shown in figure \ref{fig:safe_zone_preprocess}(a), assuming the time taken for planning is $\Delta t$ and the maximum linear velocity of the robot is $v_{max}$, the maximum forward distance within this time is $\Delta d = v_{max} \cdot \Delta t$. The robot's protective radius is $r_{protect}$, indicating that obstacles cannot appear in the light green area during this time. For simplicity in calculations, the region represented by the solid red line is considered the safe zone. The radius of sector BOC is $r_{safe} = r_{protect}+v_{max}\Delta t$, with a central angle of $\theta_{sec} = 2arctan(r_{protect} / \Delta d)$. $OA=r_{safe}\cdot sin(\theta_{sec} / 2)$ Thus, the safe zone can be represented as equation \ref{eqn:safe_dis}, where $\theta$ represents the angle between a certain direction and $OO'$, with a counterclockwise direction as the positive direction:

\begin{equation}  l(\theta) = \left\{
    \begin{array}{lc}
    r_{safe} & if \quad abs(\theta)<0.5\theta_{sec} \\
    r_{safe}\cdot sin(\theta_{sec} / 2) / sin(|\theta|) & if \quad 0.5\theta_{sec}<|\theta| \\
    \end{array}
    \right.
    \label{eqn:safe_dis}
\end{equation}

Thus, the maximum speed at which the robot advances in the direction of $OO'$ can be expressed as equation \ref{eqn:safe_velocity}, where  $v_{max}$ and $a_{max}$ are the maximum velocity and acceleration of the robot respectively.
\begin{equation}
    v = min\{ \mathop{\min}_{\theta} \sqrt{2\cdot a_{max} (d(\theta)-l(\theta))} / cos(\theta), v_{max}\}
    \label{eqn:safe_velocity}
\end{equation}

\subsubsection{\textbf{Speed Output of Each Mode}}
All robots start in a standby mode. When the target direction  $\theta_{goal}$ is received, the robot enters mode \textbf{A}. At this point, the output speed is 0, and the angular velocity $\omega = k\cdot\theta_{goal}$, where k is a tunable parameter. The robot starts to change its orientation until the target is in front of it: $abs(\theta_{goal})<\theta_0$, where $\theta_0$ is the allowable error angle determined by factors such as target angle localization error. Entering mode \textbf{B}, the robot starts moving towards the target. In this process, the planner selects safe moving direction with a small angle relative to  the goal direction in the safe area and outputs the corresponding velocity $v_{\theta_safe}$ according to equation \ref{eqn:safe_velocity} and angular velocity $\omega = k\cdot\theta$.  When there is no safe angle that satisfies $abs(\theta)< \pi / 2 - \theta_{1}$, where $\theta_1$ is a margin angle that enables the safe velocity is greater than zero to enhance movement efficiency,  it indicates the presence of an obstacle in front of the robot, reaching a key point for turning decision.At this point, the robot determines its turning direction by integrating all perceptual information using optimizer, i.e, graph neural networks. If  the optimizer recommends a left turn, the robot enters mode \textbf{C}; otherwise, it enters mode \textbf{D}. In these two modes, the robot turns in place until there exist safe zone with its midline angle satisfying $\theta < \pi/2 - \theta_{1}$ , then the robot starts circumventing the obstacle, entering mode \textbf{E}  or mode \textbf{F}. During the circumvention process, the robot's speed and angular velocity are determined by the safe zone in front of the robot. When the obstacle and the target are on opposite sides of the robot, it has successfully bypassed the obstacle. At this point, it returns to mode \textbf{A}, starts turning towards the target and moving towards it. This process repeats until the robot reaches a location where the target is detected within sight and the distance is less than $r_{goal}$.

\subsubsection{\textbf{Collision Avoidance}}
Priority-based collision avoidance strategy is adopted  for safety. When a robot perceives that a neighboring robot is approaching and a collision is possible, the lower priority robot chooses to wait while the higher priority robot treats the lower priority robot as a static obstacle. Once the higher priority robot moves to a distance greater than the safety distance from the lower priority robot, the lower priority robot stops waiting and continue to execute its task.

\subsection{Network Architecture for Top-level Intelligent Search Direction Selection}
When approaching a target and encountering obstacles, a robot must determine the best direction to circumvent the obstacles based on its perception of the surrounding environment. The conventional method of choosing a fixed direction yields only a $50\%$ probability of selecting the shorter path between two routes. To address this, we designed a graph neural network architecture that utilizes perceptual data from both the target robot and its teammates, enabling the selection of a more optimal direction for bypassing obstacles and enhancing the fundamental planner.

\begin{figure}[t]
\centering
\includegraphics[width=3.4in]{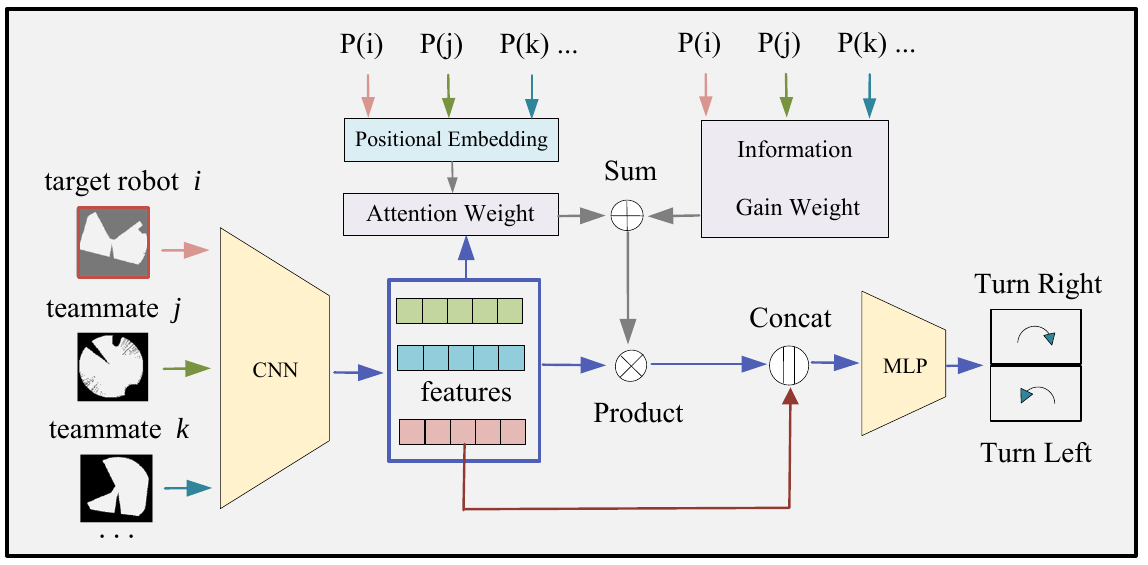}
\caption{Distributed Model Architecture, which consists of: a CNN-based feature encoder, GNN layer to fuse perceptual features with positional information, and a MLP-based direction mapper.}
\label{fig:architecture}
\end{figure}

\subsubsection{\textbf{Feature Encoder}}. For each robot, a Convolutional Neural Network (CNN) is employed to extract informative features $f\in R^F$  from rotated local maps, and then transmit these features to neighboring robots.  In this study, a mini VGG architecture\cite{simonyan2014VGG} is utilized to extract features from the local map. The mini VGG consists of a sequence of Conv2d-BatchNorm2d-ReLU-MaxPool2d and Conv2d-BatchNorm2d-ReLU blocks repeated thrice. Subsequently, a fully connected layer is employed to map the flattened CNN features into an F-dimensional vector. In this article, we always keep F=128. Using this feature extractor, we obtain the compressed feature of the two-dimensional spatial environment surrounding each robot, as detected by radar, and transmit these characteristics to nearby communicable teammates when necessary.
\begin{equation}  
    f_i = Encoder(Input_i)
    \label{eqn:encoder}
\end{equation}

\subsubsection{\textbf{Graph Attention with Information Gain Weight (GIWT)}} 
When integrating the local environmental information gathered by the target robot and its teammates, the impact of the perceived data from robots in different positions on the decision-making process varies. For example, if there is an obstacle directly in front of the robot, the information from teammates behind the robot is more influential than that from teammates in front of it . Even in the same direction, those teammates farther provide more valuable information and have a greater impact on the decision-making process.Building on this characteristic, this paper utilizes a graph attention mechanism that incorporates positional features, which enables the network to learn different weights based on the specific positions of teammates from expert data.

First, we divide the perception area of the target robot into a $7\times7$ grid, as shown by the red grid in Figure \ref{fig:safe_zone_preprocess}, and assign numbers from 0 to 48 to each small unit, with the robot itself located in grid number 24. Through the preprocessing stage, the positional identifiers of teammates can be obtained. These identifiers are then utilized in a learnable positional embedding, which allows the model to adjust the weights of the positional encodings during the training process, thereby adaptively finding the optimal encoding parameter. We combine positional encoding with attention score computation. By integrating node features with positional encoding, the model's sensitivity to node positions is enhanced, thereby improving information propagation within the graph.
\begin{equation}  
    PE(i) = PositionalEmbedding(P(i)) \in R^{F'}
    \label{eqn:positionalembedding}
\end{equation}

To compute the weights of each robot node, the positional encodings and sensory features are first combined with a linear transformation to better integrate the features before computing the weights.Then the node features integrated with positional information are used to compute the weight coefficients between nodes.
\begin{equation}  
    PF(i) =  W\cdot f_i + PE(i)
    \label{eqn:attention_weight}
\end{equation}
\begin{equation}  
    \alpha(i,j) = \frac{exp(LeakyReLU(PF(i),PF(j)))}{\sum_{k\in{i}\cup{N(i)}} exp(LeakyReLU(PF(i),PF(j))}
    \label{eqn:attention_weight}
\end{equation}

Next, information gain weight is introduced, which is determined based on the increase in the perceived area of neighboring robots relative to the central robot, to represent the extent to which the information gain obtained by the robot at different distances affects the decision-making of the target robot. Let the robot's perception area be denoted as S. The additional perceived area of neighboring robot j relative to central robot i can be represented as $S_j\backslash S_i$, where $(S_j\backslash S_i) \bigcup (S_j\bigcap S_i) = S_j,(S_j\backslash S_i) \bigcap (S_j\bigcap S_i) = \emptyset$ . If the distance between the neighboring robot and the central robot is denoted as $r$, let $q=r/R_{FOV}$, then the proportion of the additional area compared to $S_i$ can be expressed as equation \ref{eqn:info_gain_weight}. Then we use the Taylor expansion, keeping terms up to the third order, and obtain equation \ref{eqn:info_gain_weight_taylor}. 
\begin{equation}  
    \beta(i,j) =  1-2arccos(q)/\pi+2q \sqrt{1-q^2}/ \pi 
    \label{eqn:info_gain_weight}
\end{equation}
\begin{equation}  
    \beta(i,j) \approx  4q/\pi-2q^3/ 3\pi 
    \label{eqn:info_gain_weight_taylor}
\end{equation}
Finally, the aggregation process of the graph neural network with positional weights is represented by Equation \ref{eqn:aggregate_func}, and the fused perceptual feature output will be mapped to the selection of left and right turning actions by an MLP network.
\begin{equation}  
    h = \sigma (\sum_{k\in{i}\cup{N(i)}} {(\alpha(i,j)+\beta(i,j))W\cdot PF_k } )
    \label{eqn:aggregate_func}
\end{equation}

\subsubsection{\textbf{Direction Mapper}} After the fusion of perceptual information, the fused feature is mapped to the output probabilities of two different directions using a Multi-Layer Perceptron (MLP) network. This network consists of two linear layers and one non-linear activation layer, with an output dimension of 2. The output probabilities in the two directions are normalized probabilities obtained by applying softmax-transformation to the neural network outputs.

\subsection{Expert Data Generation and Model Training}
\subsubsection{\textbf{Expert Data Generation}}
Although directly running the DHbug algorithm to identify key points and record expert data can produce expert data consistent with actual applications, this method is very time-consuming, the speed of data generation is too slow. Therefore we designed an expert data generation scheme for the training of the optimizer.

In order to cover a variety of obstacle types, we generate maps with different types of obstacles, randomly placing rectangular prisms of various sizes and aspect ratios, as well as cylinders with different base radii, and allowing the obstacles to overlap. The initial positions of each robot are randomly distributed in the blank areas of the map, and a random target is set for each robot. Each robot orients its front towards the target and then the radar data, absolute position of each robot and the target are recorded. Finally,the directions taken by the shortest path planned using the $A\star $ algorithm on the global map are used as data labels. In most cases, there are no obstacles directly in front of the randomly distributed robots, so only the data of robots with obstacles in the safe area directly in front are selected as expert data. When using the raw data generated in ROS, we transform them into a relative coordinate system to simulate real scenarios. As in Figure \ref{fig:safe_zone_preprocess}, the radar data of each robot will be transformed with the direction of the line connecting the central robot and its target as the positive direction of the coordinate axes.

\subsubsection{\textbf{Learning from Expert Data}}
During the training process, the expert dataset is divided into training, testing, and validation datasets in the ratio of 3:1:1. The training goal is to learn a mapping function $M(\cdot)$ that minimizes the discrepancy between the model output and the ground truth turning direction labels $Y\in\{0,1 \}$ from the expert dataset. Here, the input $X$ comprises radar and positioning data from the central robot and its neighboring teammates. We utilize cross-entropy loss $\mathcal{L}(\cdot)$ as the objective function for training, with the model's parameters  $\theta$ being trainable.

 \begin{equation}
 \hat{\theta} = argmin_{\theta} \mathcal{L}_{\theta}(M(X),Y)
 \label{eqn:lossfunc}
 \end{equation}

\section{Evaluation}
\subsection{Experiment Setup}
\subsubsection{\textbf{Expert Data}} We randomly generates maps with different obstacle distributions. The obstacles in the maps consist of a mix of cylindrical obstacles with varying base radii and rectangular obstacles with different aspect ratios and sizes, allowing for interconnection and overlapping between obstacles to create more complex distribution types. We use a fixed map size of $20\times20$ and varies the obstacle sizes to achieve different proportions between obstacles and the map. The range of sizes for a single edge of the rectangular obstacles is 0.5-3m, with random orientations, while the base radius of the cylindrical obstacles is randomly generated within the range of 0.5-3m. During the generation process, maps with the following types of obstacles are created: \textbf{(map A)} only rectangular obstacles, \textbf{(map B)} only cylindrical obstacles, and \textbf{(map C)} random mixtures of rectangular and cylindrical obstacles. The total number of obstacles is randomly generated between 10 and 30. To ensure that multiple robots have different graph structures, 15 individual robots are randomly generated in blank spaces.

\subsubsection{\textbf{Neural Network Parameters Setting}} In the experiment, the input dimension of the CNN network is 101*101 (2.5m maximum perception distance and 0.05m resolution) with an output feature dimension of 128. The output feature dimension of the graph neural network is also 128, with 3 layers of MLP and a middle layer dimension of 16.

\subsubsection{\textbf{Robot and Task Setting}}
We designed two different types of tasks. In the first type of task(\textbf{Task I}), the robot's starting and target positions are randomly generated from the empty spaces on the map. In the second type of task(\textbf{Task II}), the robots are lined up in a column on one side of the map and move towards the other side. Therefore, six different sets of expert data were generated by combining three different maps with two types of tasks: (\textbf{1}) map A, Task I (\textbf{2}) map B, Task I (\textbf{3}) map C, Task I (\textbf{4}) map A , Task II (\textbf{5}) map B, Task II (\textbf{6}) map C, Task II.

\subsection{Performance on Expert Datasets}
\begin{table}[t]
\renewcommand\arraystretch{1.}
\centering
\caption{Accuracy of different methods on six expert datasets.}
\label{table:expert_data_performance}
\begin{tabular}{ccccccccc}
\toprule 
\multirow{2}{*}{Methods} & \multicolumn{6}{c}{Expert Data Type} \\
\cline{2-7}
                & 1 & 2  & 3 
                &4 & 5  & 6
                \\
\midrule 
CNN       & 0.755 & 0.764 & 0.768 & 0.772 & 0.771 & 0.776 \\
GraphSAGE & 0.785  & 0.800 & 0.795 & 0.801 &  0.808 & 0.809 \\
GAT      & 0.787 & 0.792 & 0.797  & 0.810 & 0.816 & 0.814 \\
\midrule
GIWT &0.809 & 0.817& 0.816  & 0.821 & 0.820 & 0.817\\
\bottomrule
\end{tabular}
\end{table}
\subsubsection{\textbf{Performance}}
The performance of different networks on six expert datasets is shown in Table \ref{table:expert_data_performance}. Our method demonstrates better performance compared to other models and GNNs that integrate neighbor information outperforms CNN that only utilizes the perception data of the target robot. In Task II, where the robot has teammates on both left and right directions, the data pattern is simpler compared to Task I where the positions of neighboring robots are more stochastic. Therefore, the performance of various networks on the expert datasets of Task II is generally better than that of Task I. Our proposed method outperforms both CNN and classic GNN such as GAT, GraphSAGE on expert datasets, achieving accuracies of 0.809 and 0.821 on the first and fourth types of expert datasets, respectively.

\subsubsection{\textbf{TSNE Analysis}}

\begin{figure}[t]
	\centering  %图片全局居中
	\subfigcapskip=-5pt %设置子图与子标题之间的距离
        \subfigure[input]{
		\includegraphics[width=0.22\linewidth]{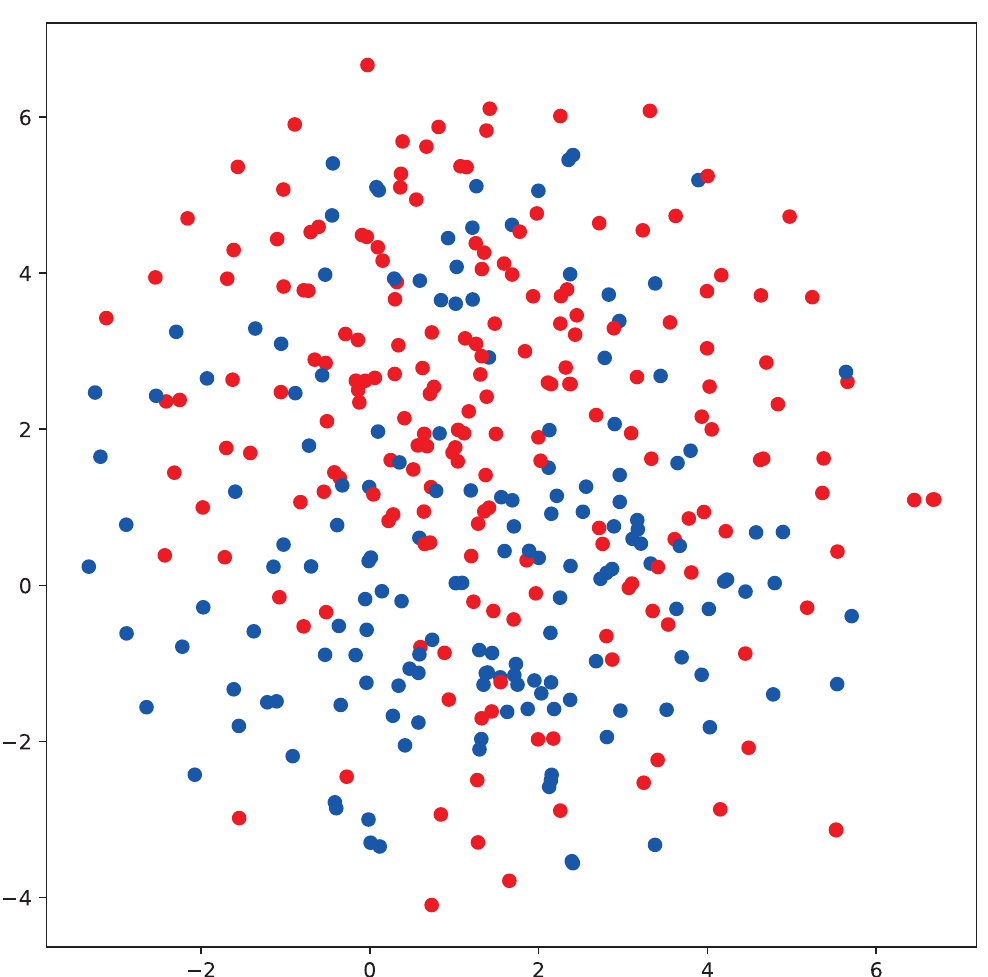}} 
	\subfigure[encoder]{
		\includegraphics[width=0.218\linewidth]{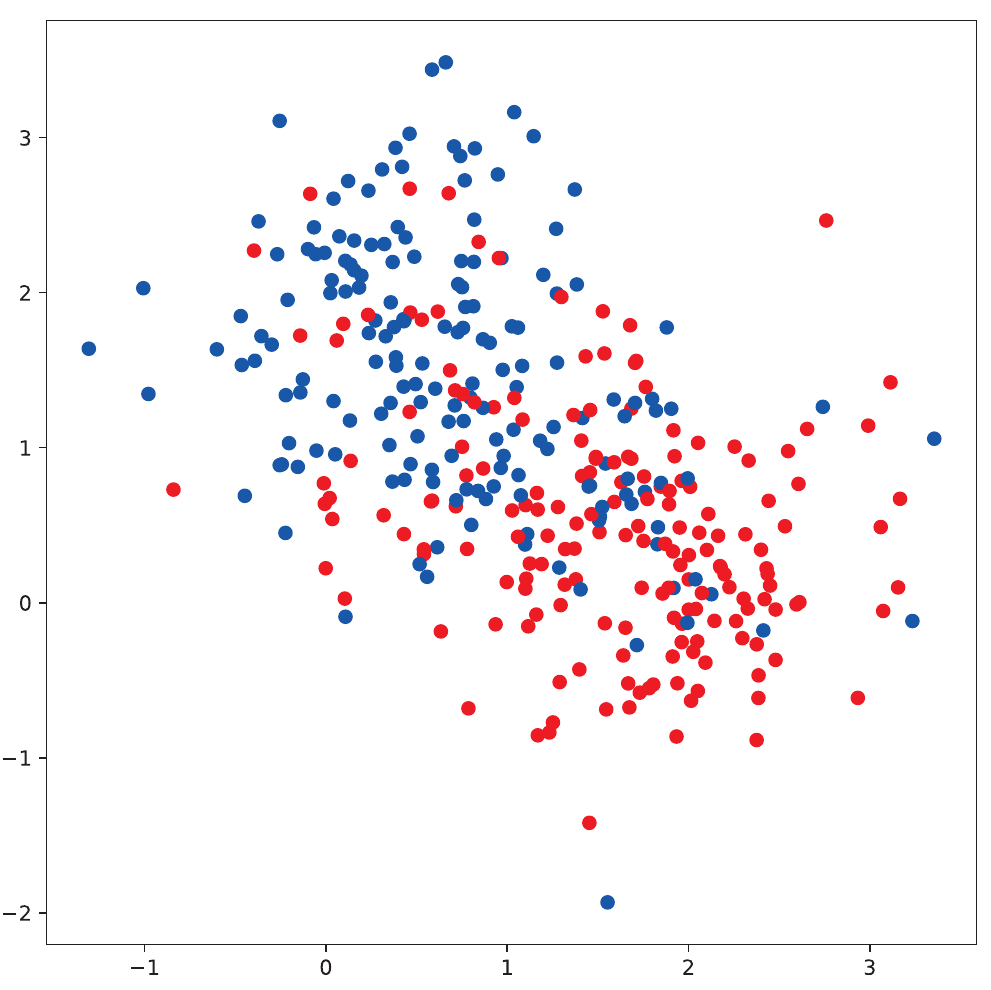}}
	\subfigure[graph layer]{
		\includegraphics[width=0.222\linewidth]{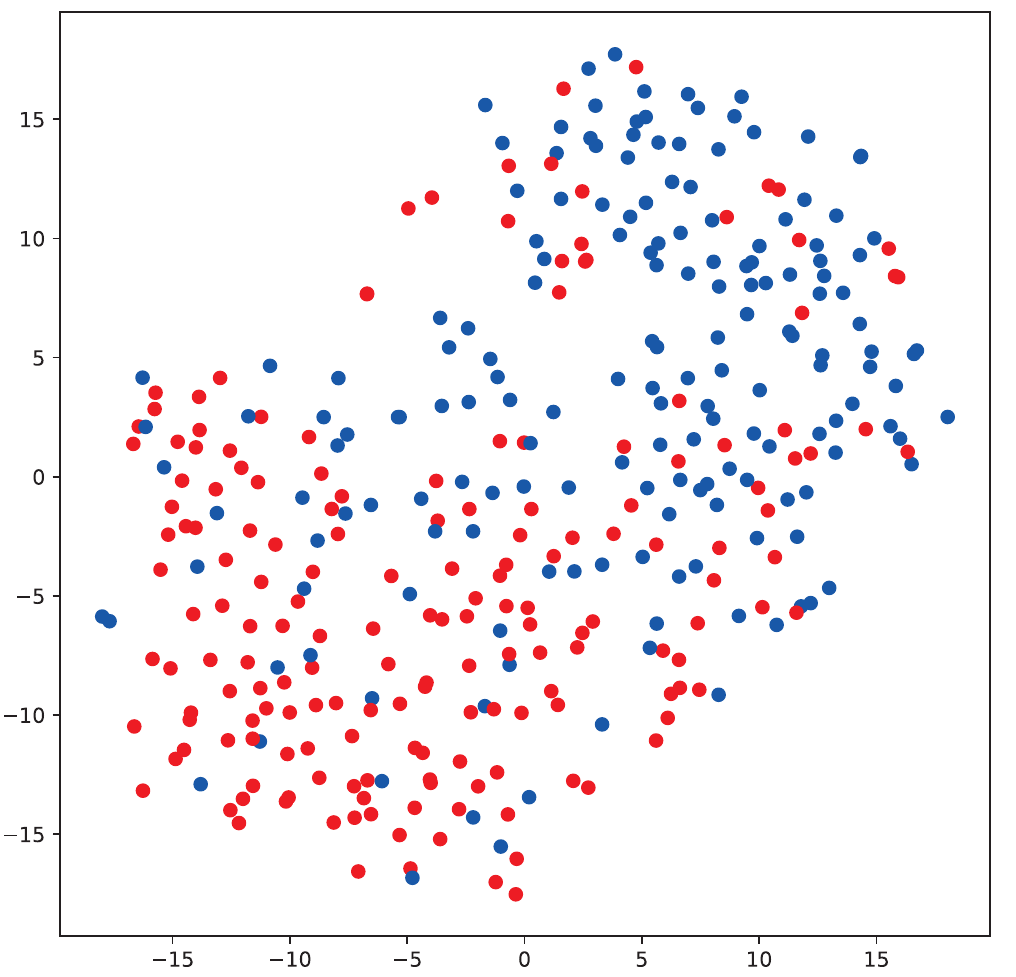}}
        \subfigure[final layer]{
		\includegraphics[width=0.22\linewidth]{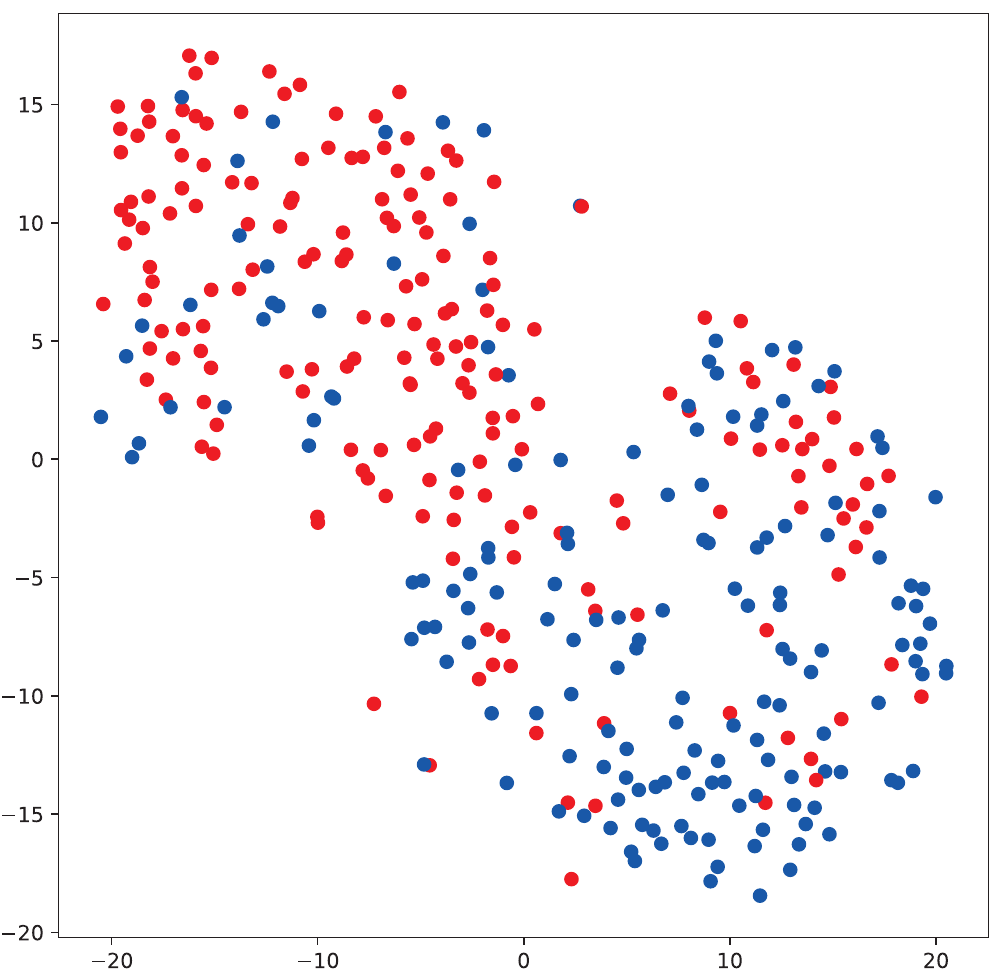}}\\
	\caption{TSNE analysis of the outputs from different layers of the trained model. Blue points indicate inputs labeled as turning left, while red points correspond to inputs labeled as turning right.}
        \label{fig:tsne}
\end{figure}

To evaluate the effectiveness of GIWT in fusing perception information from multiple robots, we use t-SNE to map the output of the trained model at different layers to points in a two-dimensional space. The points are then labeled with expert data labels of different colors, shown as Figures \ref{fig:tsne}.
It can be observed that in the output of different layers, most of the data points for turning left and right are well separated, indicating that the robots can achieve effective turning recommendations using only their own perception information or the fused information. This further validates the capability of the proposed method in combining multi-robot perception information. 

\begin{figure}[]
\centering
\includegraphics[width=1.8in]{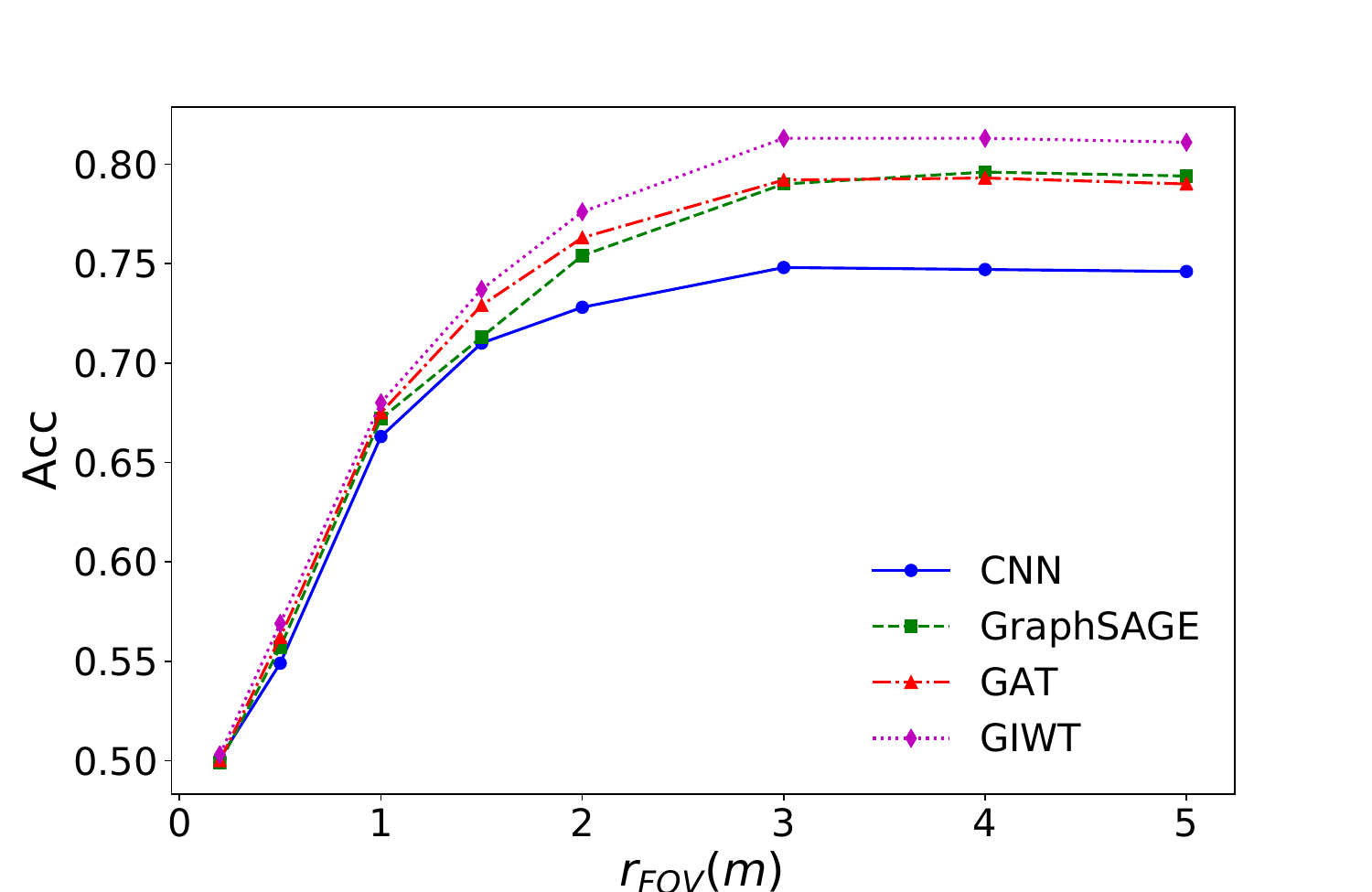}
\caption{Impact of Perception Radius on Information Fusion Performance.}
\label{fig:fov_effect}
\end{figure}

\subsubsection{\textbf{Key Parameters}} Figure  \ref{fig:fov_effect} shows the performance of different models on the entire expert dataset as the perception radius varies. It can be observed that as the perception radius increases, the accuracy of making correct steering decisions with different networksgradually improves and eventually reaches saturation. This is because, obstacles have a shielding effect on radar. Although expanding the radar radius can broaden the field of view to some extent, the blocking effect of nearby obstacles on the rear space becomes more pronounced. It can be seen that GIWT has a slightly stronger fusion effect compared to other graph neural networks.

\subsection{Path Planning Test in ROS}
\begin{table}[t]
\renewcommand\arraystretch{1.1}
\centering
\caption{Comparison of the effects of different methods optimizing the DHbug algorithm in two tasks.}
\label{table:ros_test}
\begin{tabular}{ccccccccc}
\toprule 
\multirow{2}{*}{Methods} & \multicolumn{3}{c}{Task I} & \multicolumn{3}{c}{Task II} \\
\cline{2-7}
                &{\scriptsize  $ Acc \uparrow $} & {\scriptsize $APL \downarrow$}  & {\scriptsize SR$\uparrow $} 
                &{\scriptsize  $ Acc\uparrow $} & {\scriptsize $APL \downarrow$}  & {\scriptsize SR$\uparrow $}
                \\
\midrule
{turn left} & 0.509 & 15.28 & 95.4$\%$ & 0.497 & 24.46 & 96.0$\%$  \\
CNN       & 0.760  & 14.76 & 97.4$\%$ & 0.775 &  23.68  & 96.2$\%$ \\
GraphSAGE & 0.788  & 14.24 & 97.6$\%$ & 0.805 &  23.27 & 97.8$\%$\\
GAT      & 0.793 & 14.37 & 98.2$\%$ & 0.812 & 23.13 & 97.4$\%$\\
A*         & 1.000 & 13.29 & 99.6$\%$ & 1.000 & 22.16 & 99.4$\%$ \\
\midrule
GIWT &0.813& 14.04& 98.8$\%$&0.819 & 22.85 &98.3$\%$\\
\bottomrule
\end{tabular}
\end{table}

We tested the proposed method in ROS on Task I , Task II and recorded the execution trajectories and path lengths. For each task, we randomly generated 500 different maps and had the robot start from the initial point. The task was terminated either when the robot reached the target point or when the execution time exceeded the maximum time limit(5 mins). 

\begin{figure}[]
\centering
\includegraphics[width=2.8in]{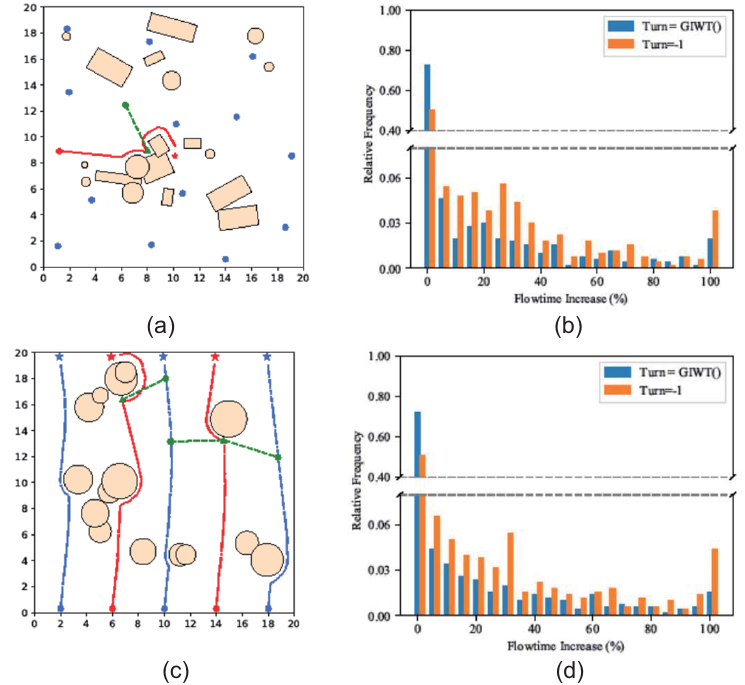}
\caption{Two cases(Figure a and c) from the ROS tests and  Flowtime Increase(Figure b and d) of the proposed method against continuous right turns. Figure a , b come from Task I,  c and d come from Task II.}
\label{fig:case_statiscs}
\end{figure}

In Figure \ref{fig:case_statiscs} (a) and \ref{fig:case_statiscs} (c) , the red dot and  the red star mark the initial position and target position of the observed robot respectively. For Tasks I, the positions of other robots remain unchanged, with blue dots representing stationary robots. In Tasks II, all robots move from their starting points to the targets ahead. The trajectories of robots optimized using neighbors' information are shown in red, while others are depicted in blue. Green dots indicate the positions of nearby teammates when the target robot makes decisions, and green dashed lines represent mutual visibility and communication between them. In Figure \ref{fig:case_statiscs} (b) and \ref{fig:case_statiscs} (d) , we compared the optimization performance of classical DHbug that always turn right  and GIWT optimized DHbug in terms of trajectory and presented a frequency histogram regarding the optimization effects.The x-axis represents Flowtime Increase, computed as $FT = (l - l^\star) / l^\star$ where $l^\star$ is the shortest path length produced by DHbug. This value indicates the percentage increase in the actual trajectory length relative to the optimal trajectory, with larger values indicating worse optimization results. The y-axis represents the frequency corresponding to each FT value. From the comparison of frequency histograms across different tasks, it is evident that the proposed method shows a significant optimization effect compared to the algorithm without collaboration.

The statistical results are shown in Table \ref{table:ros_test}, where accuracy(acc) represents the proportion of correct decisions compared to the expert data using global map information in 500 experiments, the Average Path Length (APL) indicates the average path length of the robot in 500 experiments, and the Success Rate (SR) represents the arrival rate within 5 minutes.
 From this table, it can be observed that the optimization result of the $A\star$ algorithm with global visibility serves as the upper limit of optimization for various networks. Compared to the classical DHbug algorithm, the average path length is reduced by approximately $13\%$ in Task I and $9.4\%$ in Task 4. Our proposed method achieves a reduction of about $8.2\%$ in Task I and $6.6\%$ in Task II. Furthermore, the expert algorithm has an arrival rate of approximately $99.6\%$ in Task I and $99.4\%$ in Task II within 5 minutes, while our proposed method achieves about $98.8\%$ and $98.3\%$, showing clear superiority over other methods.

\begin{figure}
	\centering  %图片全局居中
	\subfigcapskip=-5pt %设置子图与子标题之间的距离
        \subfigure[turn right]{
		\includegraphics[width=0.3\linewidth]{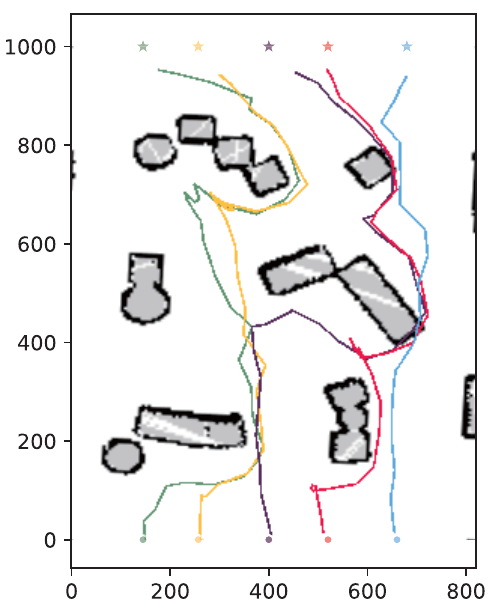}} 
	\subfigure[CNN]{
		\includegraphics[width=0.302\linewidth]{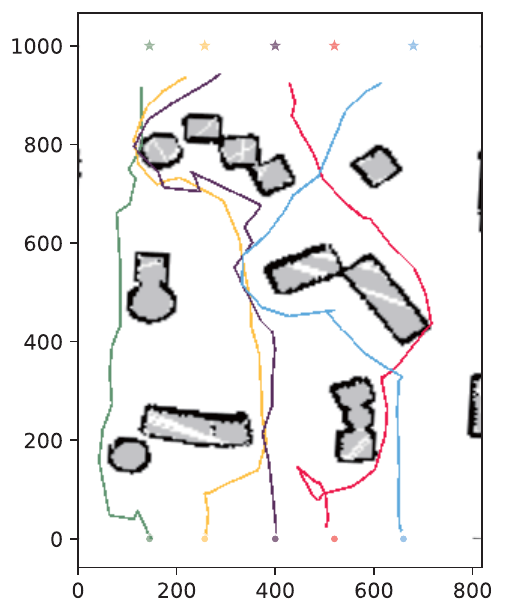}}
  	\subfigure[GIWT]{
		\includegraphics[width=0.3\linewidth]{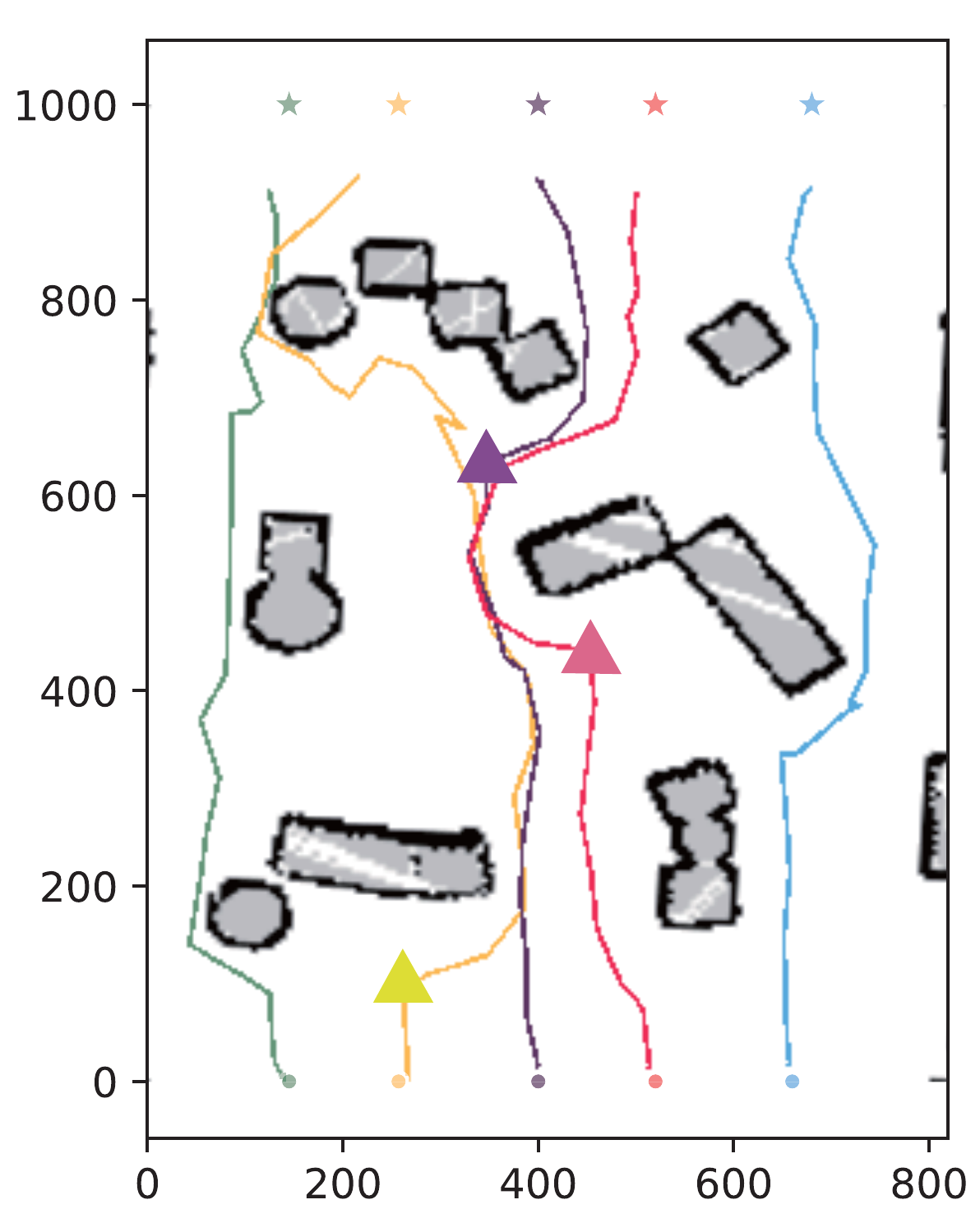}}
        \subfigure[ key point 1 \qquad \qquad  (e) key point 2 \qquad  \qquad (f) key point 3]{
		\includegraphics[width=0.95\linewidth]{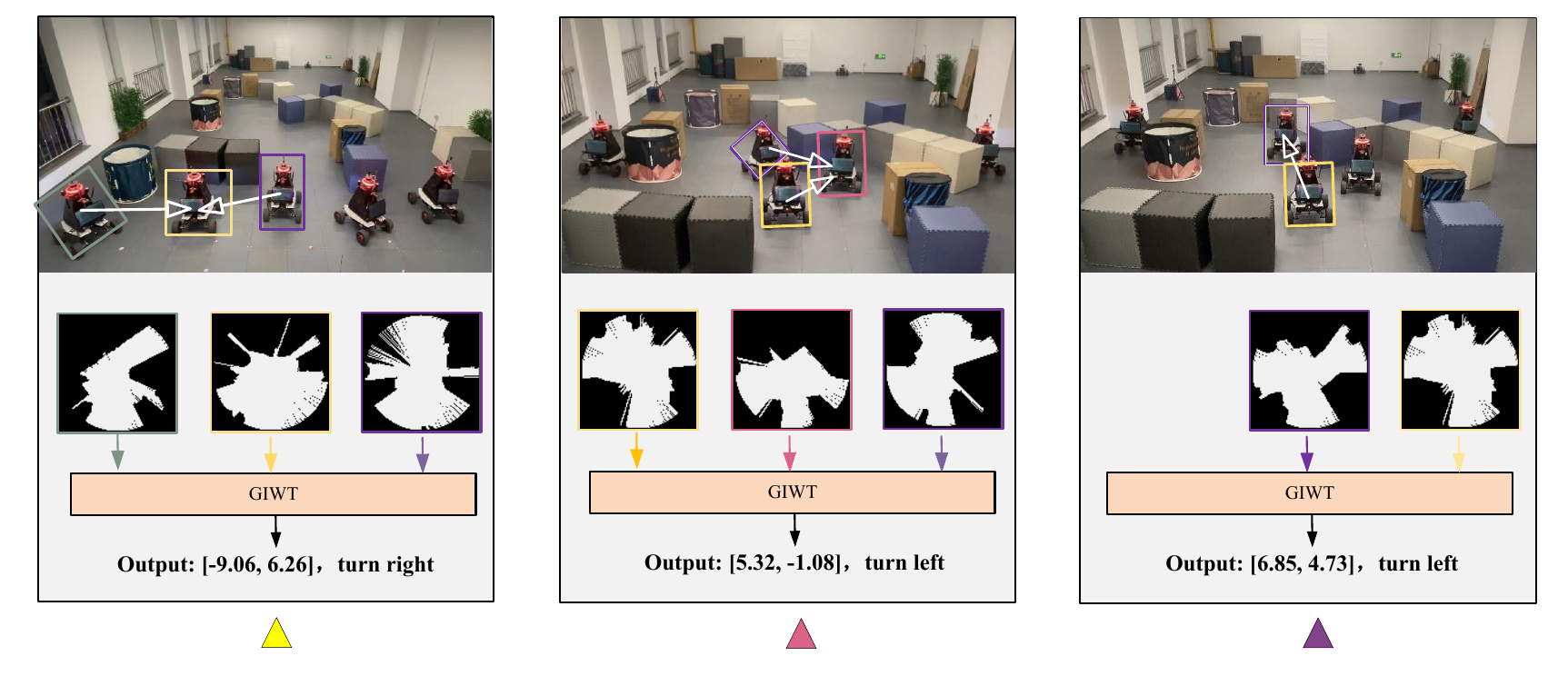}}
    \caption{Comparison of Execution Trajectories (a)-(c) and Analysis of Decision Points (d)-(f) for the Proposed Method in a Practical Application Scenario.} 
        \label{fig:giwt_realword}
\end{figure}

\subsection{Real-world Experiments}
We also conducted physical experiments on the multi-robot experimental platform established in our laboratory. In an 8m x 10m area, the absolute positions of five omnidirectional wheeled robots were determined using UWB devices placed at the four corners of the field. This information was used to record trajectories from observer insight. The robots identified their teammates and determined relative positions using cameras and UWB devices. Each robot uses its radar with a 2.5m detection radius to detect the local environment. We tested the robot traversal tasks on 10 randomly generated maps.Table \ref{table:physical_test} presents detailed data from the 5 maps along with the average path length of all experiments. It can be seen that after multiple experiments, the average path lengths for fixed left turns(Turn=1), fixed right turns(Turn=-1), or random direction selection are relatively close, while the search direction selection based on GIWT achieves better optimization results compared to the method solely based on CNN. This validates the advantage of the proposed method in analyzing perceptual data and making real-time decisions with multiple robots compared to a single robot.

Figure \ref{fig:giwt_realword} shows the paths of different methods in case 2 of Table \ref{table:physical_test}, as well as real scenarios at key decision points with GIWT.
In Figure \ref{fig:giwt_realword}(a) to (c), the robots numbered 1 through 5 are shown from left to right, with bottom circles representing the initial positions and top stars indicating the target positions of the robots. Figures \ref{fig:giwt_realword} (d), (e), and (f) correspond to the decision points and robot statuses for the three triangular symbols in Figure \ref{fig:giwt_realword} (c). The white arrows indicate data transmission from communicable neighboring teammates during decision-making. At the turning position of Robot 2 (yellow triangle and Figure \ref{fig:giwt_realword}d), the radar detects obstacles ahead with similar shapes on both sides. Although the CNN output [2.75, 2.93] suggested a right turn, this position is susceptible to disturbances from minor changes in the perceived data. In contrast, the GIWT network, which incorporates nearby robot perception data, provides a more reliable decision to turn right. At the turning position of Robot 4(the pink triangle and Figure \ref{fig:giwt_realword}e), the traditional DHbug algorithm suggests a right turn, which would require the robot to navigate around most of the obstacle's right edge, while GIWT recommends a left turn, effectively utilizing the observed open space on the left. At the turning position of Robot 3(purple triangle and Figure \ref{fig:giwt_realword}(f)), although there is a visible and communicable Robot 2 nearby, the latter is positioned far behind Robot 3, resulting in a lower weight of 0.07 during information fusion. Therefore, Robot 3 primarily relies on its own information for decision-making.

\begin{table}[t]
\renewcommand\arraystretch{1.1}
\centering
\caption{Average Path Length across Ten Maps in Physical Experiment.}
\label{table:physical_test}
\begin{tabular}{ccccccccc}
\toprule 

case  & turn left & turn right  & turn randomly & CNN & GIWT   \\
 \midrule
  1  & 14.83  & 13.52   & \textbf{12.96}  &  13.68 &  13.19\\
   2 (Fig. \ref{fig:giwt_realword})   & 12.55  & 14.21   &  14.01 & 12.28 & \textbf{11.03}\\

 3   &  11.49  & 12.07   &  11.73 &    \textbf{11.01} & 11.39\\
 4   & 13.94  & 12.76 & 12.89 & 12.07 & \textbf{11.90}\\
 ...   &  ... & ...   & ...  & ...& ... \\
10  &  \textbf{12.36} &  14.76  &  13.91 & 12.77 &  12.59\\
\midrule  %添加表格中横线

average  &  13.21     & 13.27 & 13.10 &  12.47 & \textbf{11.83}  \\

\bottomrule
\end{tabular}
\end{table}

\subsection{Discussion}
The experimental results of expert that utilizes global information demonstrate the need for further research to enhance the optimization results' upper limit, it is also noted that the optimality under local information conditions does not guarantee optimality under global information conditions. This, to some extent, affects the actual effectiveness of the method and leads to certain discrepancies between its performance and that of expert algorithms.

\section{Conclusion}

In this paper, we introduce a hierarchical collaborative path planning method for multi-robots in unknown environments. The approach leverages the DHbug algorithm, which converges mathematically, as a fundamental planner to guide robots to navigate towards targets while avoiding obstacles, using radar data to output linear and angular velocities. In situations where obstacles obstruct the path directly ahead, we propose GIWT, a Graph
Attention Architecture with Information Gain Weight, to integrate radar perception data from the central robot and its adjacent teammates, selecting shorter paths to circumvent obstacles. To ensure the trained network can effectively support the DHbug algorithm and exhibit strong generalization, an expert data generation scheme was carefully devised. 
The experimental results demonstrate the advantage of the proposed method to optimize the classical DHbug path planning algorithm in analyzing and utilizing local environmental information during state transitions in complex scenarios. 
In the future, we will further enhance the collaborative planning performance of robots by leveraging visual perception capabilities.

\bibliographystyle{IEEEtran}
\bibliography{reference}

\end{document}